# Retinex filtering of foggy images: generation of a bulk set with selection and ranking

R. Marazzato[1] and A.C. Sparavigna[2]

1 Department of Control and Computer Engineering, Politecnico di Torino, Italy
2 Department of Applied Science and Technology, Politecnico di Torino, Italy

**Abstract:** In this paper we are proposing the use of GIMP Retinex, a filter of the GNU Image Manipulation Program, for enhancing foggy images. This filter involves adjusting four different parameters to find the output image which has to be preferred according to some specific purposes. Aiming to obtain a processing, which is able of choosing automatically the best image from a given set, we are proposing a method for the generation a bulk set of GIMP Retinex filtered images and a preliminary approach for selecting and ranking them.

**Keywords:** GIMP Retinex, GIMP, Image processing, Bulk generation of images, Bulk manipulation of images.

## Introduction

The Retinex methods for filtering images – among them we have that of the GNU Image Manipulation Program – had been developed to solve some experimental observations. These experiments show the fact that it is easy to find discrepancies between an image we have recorded by a camera and the real scene we have observed. Due to some peculiar features of the human vision, concerning colors, brightness and contrast of a scene, we are able to see details both in the shadows and in the nearby illuminated areas, whereas a photograph of the same scene shows either the shadows as too dark or the bright areas as overexposed [1-3]. Inspired by the human vision biological mechanism to adapt itself to these conditions, several algorithms of image processing had been developed, which are known as Retinex methods. The first one was conceived by Edwin H. Land, an American scientist and inventor, best known as co-founder of Polaroid Corporation [4-8]. As explained in Ref.1, through the years, Land evolved several models until his last model proposed in 1986. The term "retinex" was coined by Land himself, combining the words "retina" and "cortex", to indicate the results of his researches, that human color perception involves all levels of vision processes, from the retina to the cerebral cortex.

Several Retinex approaches exist [3,9]: the Single-Scale Retinex (SSR), the Multi-Scale Retinex (MSR), and, for color images, the Multi-Scale Retinex with Colour Restoration (MSRCR). GIMP Retinex is a freely available tool of this last family, developed by Fabien Pelisson [10]. The resulting image of this filter can be adjusted selecting different levels, scales and dynamics. In this tool, there are three "levels": uniform, which tends to treat both low and high intensity areas fairly, low, that "flares up" the lower intensity areas on the image, and high that tends to "bury" the lower intensity areas in favor of a better rendering of the clearer areas of the image. The "scale" determines the depth of the Retinex scale. Minimum value is 16, a value providing gross, unrefined filtering. Maximum value is 250. The default value is 240. A "scale division" determines the number of iterations in the multiscale Retinex filter. The minimum required, and the recommended value is three. The "dynamic" slider allows adjusting color saturation contamination around the new average color (default value is 1,2).

In fact, if we are planning the use of GIMP Retinex in an automatic approach for computer vision, we have the problem of managing the huge set of images which we can obtain from it when selecting its different parameters. An important and remarkable fact is that the images obtained after the Retinex filtering have quite different histograms. Therefore, as discussed in [11], a further image bi-level thresholding of them can produce different results: this is important for the family of images, which are the subject of this paper, that is, the images which are obtained when there are foggy weather conditions. The foggy images have poor contrast and strong loss of color characteristics. In the Figure 1 we can see an example of Retinex filtering and of its effect on a bi-level thresholding [11]: a low level filtering improves the visibility even in the bi-level black and white images. This fact stimulated the present research on the use of a bulk set of filtered images.

Let us note that several methods had been proposed to improve foggy images (see [12-14] and references therein), and, among them, several Retinex algorithms had been proposed too [15-18]. Here we discuss the use of GIMP Retinex. Aiming to obtain a processing which allows to automatically choose the best image from a given set, we are proposing to generate a bulk set of GIMP Retinex filtered images and to select and rank them. In this paper, we are giving a preliminary analysis of this method, to test the most effective approach for an automatic determination of the best choice.

**Generation of the bulk transformed set**

Let us start start from a low quality image $B$ and transform it into a set of preprocessed images $T$. Using Retinex with all different values of its input parameters ($s,n,d,l$) in their range $\Omega$, we obtain the first part of $T$. The second set we join to it is the threshold-sharpened output of the first:

$$(1) \quad T = \{Rx_{s,n,d,l}(B), (s,n,d,l) \in \Omega\} \cup \{Th_t(Rx_{s,n,d,l}(B)), (s,n,d,l) \in \Omega\}$$

where: $Rx$ is the Retinex transform operator, $B$ is the starting image, $s$, $n$, $d$ and $l$, are the *scale, scale division, dynamics* and *level* parameters respectively. $\Omega$ is the range set for the quadruple of the Retinex parameters, $Th$ is the *Thresholding operator* and $t$ the *threshold* value of $Th$ operator.

Each of the four parameters $s,n,d$ and $l$, is allowed to take values in its specific domain $D_s$, $D_n$, $D_d$, $D_l$, as specified by the data sheet of the Retinex plugin, but we restricted them to a small discrete subset, in order to obtain a reasonable number of output images:

$$(2) \quad \begin{aligned} s \in \Omega_s &= \{s_i\} \subseteq D_s \\ n \in \Omega_n &= \{n_i\} \subseteq D_n \\ d \in \Omega_d &= \{d_i\} \subseteq D_d \\ l \in \Omega_l &= \{l_i\} \subseteq D_l \end{aligned}$$

so $\Omega = \Omega_s \times \Omega_n \times \Omega_d \times \Omega_l$.

In the practice, we developed a simple script in Script-Fu including the five loops used to generate both $\Omega$ and the boolean control for the thresholding, and a call to the subroutines invoking both transforms. In the Figure 2, three snapshots of parts of the generated bulk of images is shown.

**High quality set selection and ranking: a proposal**

Not all images we generated are "good": some are even "worse" than the original. At this point we need a criterion to
- discriminate between acceptable and rejectable images
- rank acceptable images, in order to find the best one(s).

Some relevant issues about the quality of the processed images are the following:
- no solid nor near-solid colored areas must occur in a good image
- the detected details must be enhanced only where they really appear and not randomly in the whole image

- the detail must significantly increase in at least some area of the image.

The last condition allows us also to rank our results by the detail increase rate, if an appropriate index is chosen. We propose the following quantitative implementation of this criterion.

For each image $P \in T$:

- We consider a scalar value, such as the brightness or a specific channel (R, G, or B), for each pixel of the image under analysis.
- We split the whole image into a partition of N equally spaced areas $\{a_{P,i} : i \in \{1,...,N\}\}$, for instance into N = 5 horizontal stripes (see Figure 3) or into a 3 x 3 rectangular lattice.
- Each area contains a subset of the pixels of the whole image. We compute the variance of their values as Absolute Area Variance ($AAV_{P,i}$)

(3) $\quad AAV_{P,i} := Var(a_{P,i})$

then we refer this value to the overall variance of the image P, obtaining the Relative Area Variance ($RAV_{P,i}$)

(4) $\quad RAV_{P,i} := \dfrac{AAV_{P,i}}{Var(P)}$ .

- We notice that also for the original image the previous quantity, which can be written as $RAV_{B,i}$, is meaningful, so we consider the ratio of this last result over the corresponding value in the unprocessed image B, so obtaining the Variance Versus Original ($VVO_{P,i}$)

(5) $\quad VVO_{P,i} := \dfrac{RAV_{P,i}}{RAV_{B,i}}$

- Now we consider the set of the N relative variances $RAV_P := \{RAV_{P,i} : i \in \{1,...,N\}\}$ and we compute its variance; we will call it Absolute Variance of Variances ($AVV_P$)

(6) $\quad AVV_P := Var(RAV_P)$

- Again, we consider the ratio of this last result over the corresponding value in the unprocessed image B, so obtaining the Relative variance of Variances ($RVV_P$)

(7) $\quad RVV_P := \dfrac{AVV_P}{AVV_B}$

Now it is possible to translate the image quality requirements into conditions on the above defined statistical quantities.

- Solid or near solid areas would show a very low (even zero in some case) local variance; we can set a threshold under which the image is thrown into the set of "bad" images, S:

(8) $\quad \epsilon \ll 1 : (\exists i : RAV_{P,i} < \epsilon) \Rightarrow P \in S$

- An image in which the transform "detected" fake details everywhere would have a very high variance in each area, but small relative variations between all areas. In this case, the quantity to compare to a threshold is $RVV_P$. The value of such threshold τ can be 1.0 or more, so we can quantitatively state that only images with at least the same variance of variances as the original or better with some minimum improvement can be accepted.

(9) $\quad \tau \geq 1 : (RVV_P < \tau) \Rightarrow P \in S$

- The increased level of detail in some area is measured through $VVO_{P,i}$. An improvement is shown when for some area this quantity is greater than 1.0, but a "nice" improvement could lead to an even higher value. Also in this case, the comparison against a threshold μ can tell whether the image is "good".

(10) $\quad \mu \geq 1 : (\exists i : VVO_{P,i} > \mu) \Rightarrow P \notin S$

The maximum value of $VVO_{P,i}$ in the preprocessed image can also be used as a rank variable to state which image is not only "good" but also "better" than another one.

(11) $\quad max(VVO_{P1,i}) > max(VVO_{P2,i}) \Rightarrow rank(P1) \, rank(P2)$

Let us show a simple example to illustrate the previously proposed approach. Let us consider the original image (upper-left panel) in the Figure 4, and five images from the bulk set. In this same Figure, we show a table concerning the variances of the six images too. In the table, cells in orange correspond to the value of $RAV_{P,i}$, in yellow to $VVO_{P,i}$, and in light gray to $AVV_P$ (black) and $RVV_P$ (red). Note that, as previously stated, the original image is "good".

The image, given in the upper-right panel, is "bad": it is an output of the Retinex filter where the transform "detected" fake details everywhere. The two lower panels in the Figure 4 are showing "bad" images too, because they have solid or near solid areas. Therefore, according to the proposed method, only the two images in the middle are "good", and in fact, we can easily appreciate this fact. After the selection of these two images, we could "rank" them according to the value of $RVV_P$ (red). Of course, here we have just shown a preliminary work on the method: we are testing several bulk sets from other images, to test the reliability of our approach.

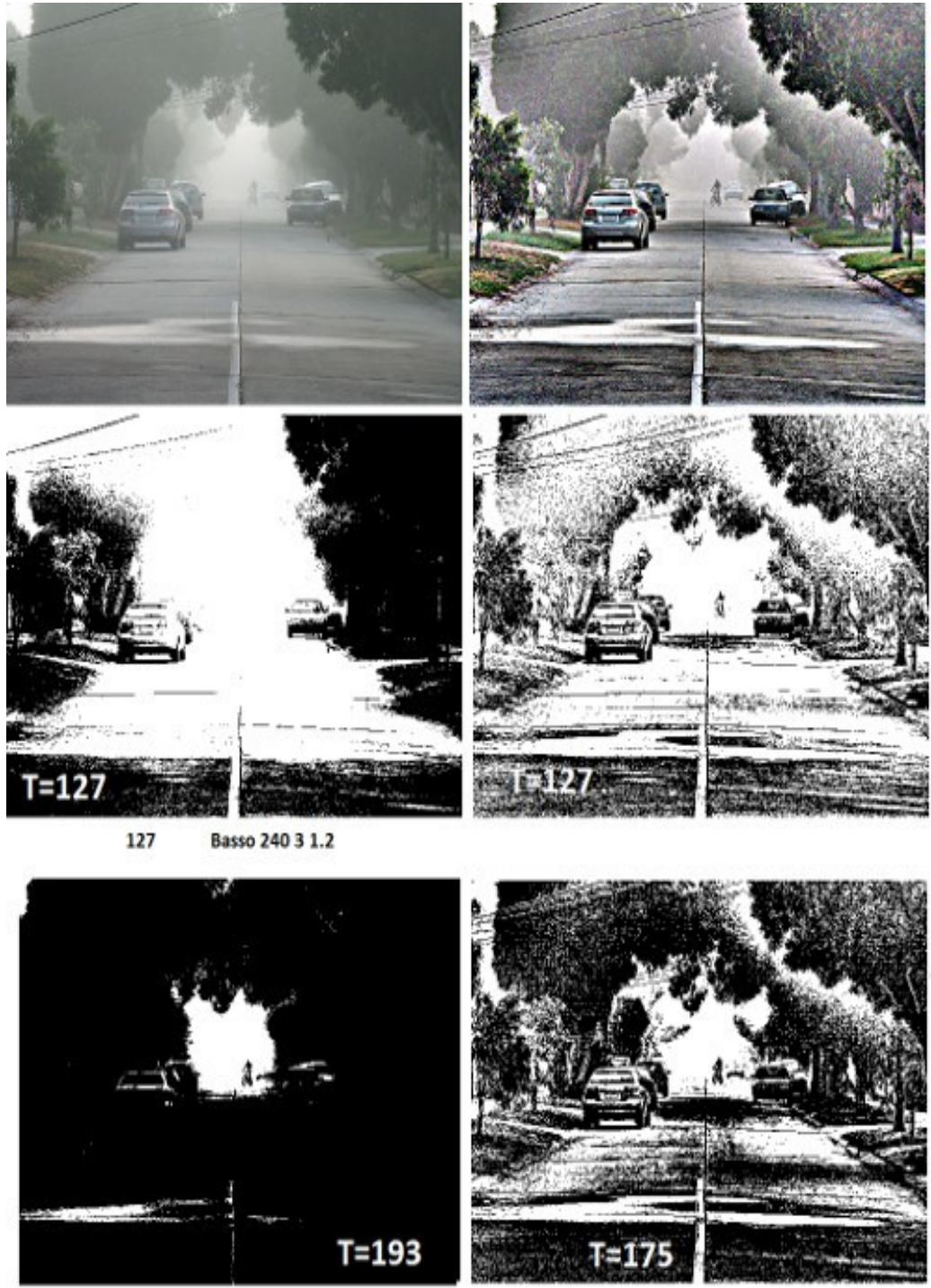

**Figure 1** - The original image in the left-upper panel, (Courtesy: Ian W. Fleggen, Wikipedia, 20880313, Foggy Street), shows a suburban street. Even a light fog is able reducing visibility, rendering the cyclist very hazy. However, a Retinex low level filtering, (with scale:240, scale division:3, dynamic:1,2), improves the visibility (right-upper panel). Let us apply a bi-level thresholding. With T=127, the original and the Retinex images appear as in the middle panels. The cyclist is visible in the Retinex image. Using the original image, to see the cyclist we have to rise the threshold to the value of 193, but in this case, we are wasting the visibility of the road. In the right-lower panel we can see a thresholding, with T= 175, of the Retinex image.

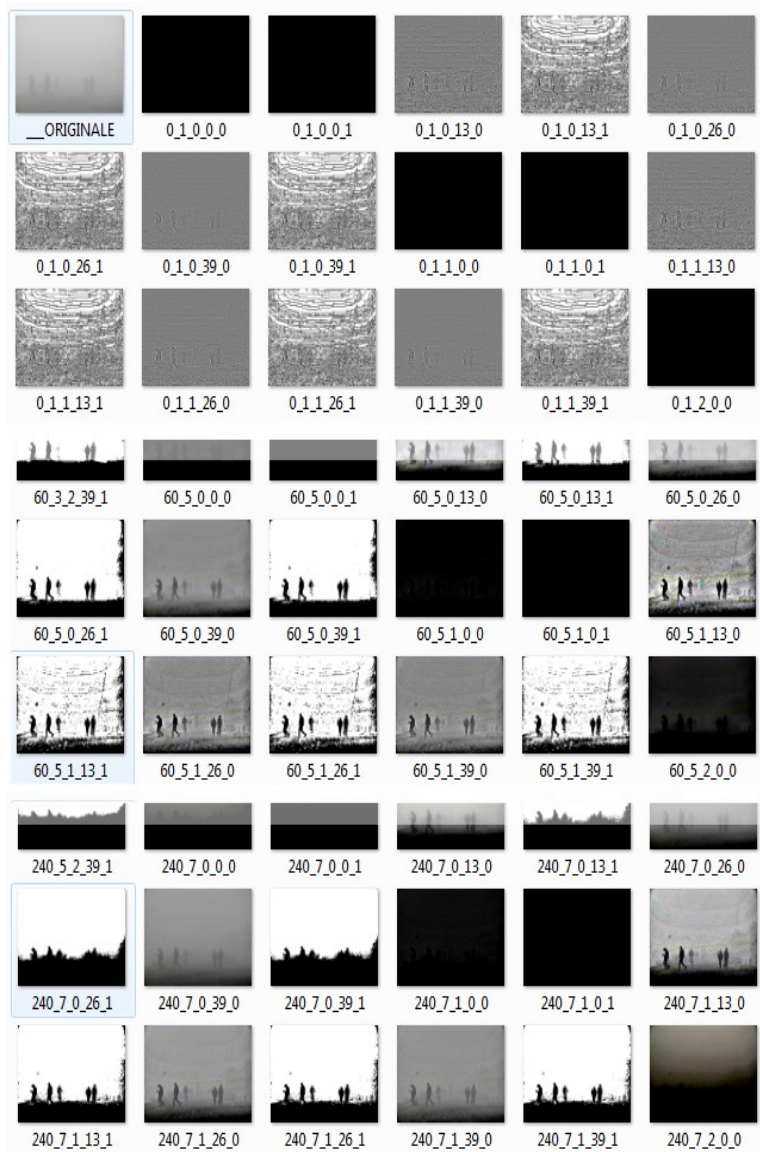

**Figure 2** - The original image in the left-upper panel. The image collects three snapshots of the bulk set.

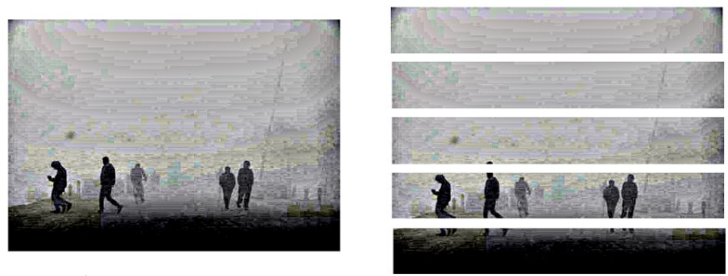

**Figure 3** – For the analysis, each image is subdivided in stripes; in our case, five stripes, as shown in the image on the right.

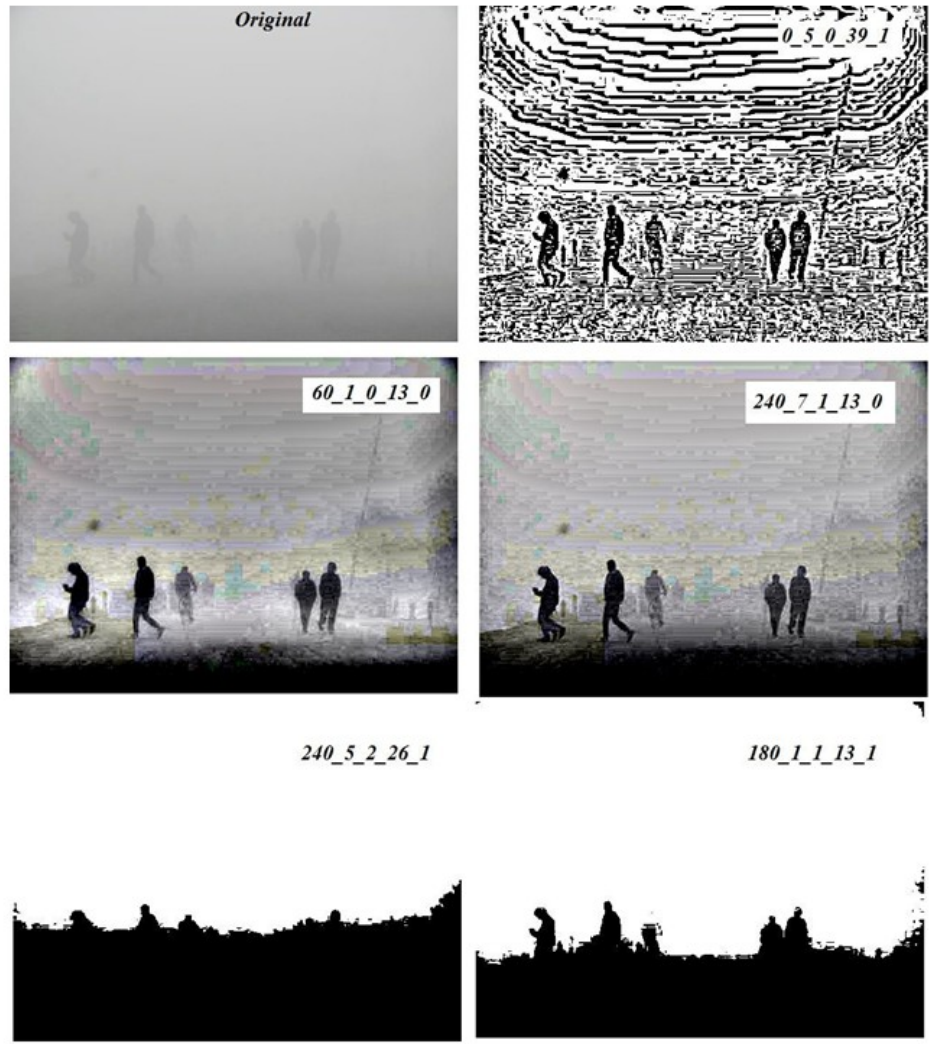

**Figure 4** – The figure shows the original foggy image and five of the images obtained after GIMP Retinex filtering. In the table, cells in orange correspond to the value of RAV$_{Pi}$, in yellow to VVO$_{Pi}$, and in light gray to AVV$_P$ (black) and RVV$_P$ (red). The image in the upper-right panel, such as the lowers which display solid colors, are "bad"; the two images in the middle are "good" according to the proposed method (see text for explanation).